\documentclass[letterpaper,10pt,conference]{ieeeconf}
\IEEEoverridecommandlockouts
\usepackage{hhline}
\usepackage{graphicx}
\usepackage{booktabs}
\usepackage{multirow}
\usepackage{amsmath}
\usepackage{amssymb}
\usepackage{array}
\usepackage[table]{xcolor}
\usepackage{pifont}
\usepackage{url}
\usepackage[nocompress]{cite}

\makeatletter
\let\NAT@parse\undefined
\makeatother
\usepackage[colorlinks=true,linkcolor=black,citecolor=black,urlcolor=blue]{hyperref}
\newcolumntype{L}[1]{>{\raggedright\arraybackslash}p{#1}}

\definecolor{yesmark}{RGB}{62,94,42}
\definecolor{partialmark}{RGB}{208,121,27}
\definecolor{nomark}{RGB}{138,138,138}
\definecolor{vlaTint}{RGB}{235,242,232}
\definecolor{wamTint}{RGB}{252,239,220}
 \definecolor{projectpink}{RGB}{190,80,125}
\newcommand{\best}[1]{\textbf{#1}}
\newcommand{\second}[1]{\underline{#1}}
\newcommand{\drop}[1]{\,\raisebox{0.25ex}[0pt][0pt]{\scalebox{0.75}{\tiny\textcolor{gray}{$\downarrow$\makebox[1.8em][r]{#1}}}}}
\newcommand{\nodrop}{\hphantom{\drop{0}}}

\newcommand{\barlabel}[1]{{\tiny\fontencoding{OT1}\fontfamily{cmr}\fontshape{it}\selectfont #1}}
\makeatletter
\newsavebox{\LIBEROsavedstrutbox}
\newcommand{\groupbar}[3]{%
  \noalign{\global\setbox\LIBEROsavedstrutbox=\copy\@arstrutbox
    \global\setbox\@arstrutbox=\hbox{\vrule height 3.8pt depth 0.5pt width 0pt}}%
  \rowcolor{#2}%
  \multicolumn{#1}{c}{\barlabel{#3}}\\[-0.5pt]%
  \noalign{\global\setbox\@arstrutbox=\copy\LIBEROsavedstrutbox}}
\makeatother
\newcommand{\markbox}[1]{\makebox[1.2em][c]{#1}}
\newcommand{\colhdr}[2]{\begin{tabular}[c]{@{}c@{}}#1\\#2\end{tabular}}
\newcommand{\cmark}{\markbox{\textcolor{yesmark}{\ding{51}}}}
\newcommand{\pmark}{\markbox{\textcolor{partialmark}{\ensuremath{\triangle}}}}
\newcommand{\xmark}{\markbox{\textcolor[HTML]{B03A2E}{\ding{55}}}}
\title{\LARGE \bf
LIBERO-VPro: Benchmarking Closed-Loop Visual Robustness of Robotic Foundation Models}
\author{%
  \authorblockN{
  Huiqiong Li$^{1}$, Zhiting Mei$^{2}$, Anirudha Majumdar$^{2}$,
  Jingjing Chen$^{3}$, Yu-Gang Jiang$^{3}$,
  Bin Zhu$^{1}$\authorrefmark{2}
  }
  \authorblockA{
  $^{1}$Singapore Management University \quad
  $^{2}$Princeton University \quad
  $^{3}$Fudan University\\
  \textbf{Correspondence:} \texttt{binzhu@smu.edu.sg}\\
  \textcolor{projectpink}{\url{https://huiqiongli.github.io/LIBERO-VPro/}}
  }
  \thanks{\authorrefmark{2}Corresponding author and project lead.}%
  }
\makeatletter
\IEEEaftertitletext{%
  \vspace{-1\baselineskip}%
  \begin{minipage}{\textwidth}%
    \centering
    \def\@captype{figure}%
    \includegraphics[width=0.95\linewidth]{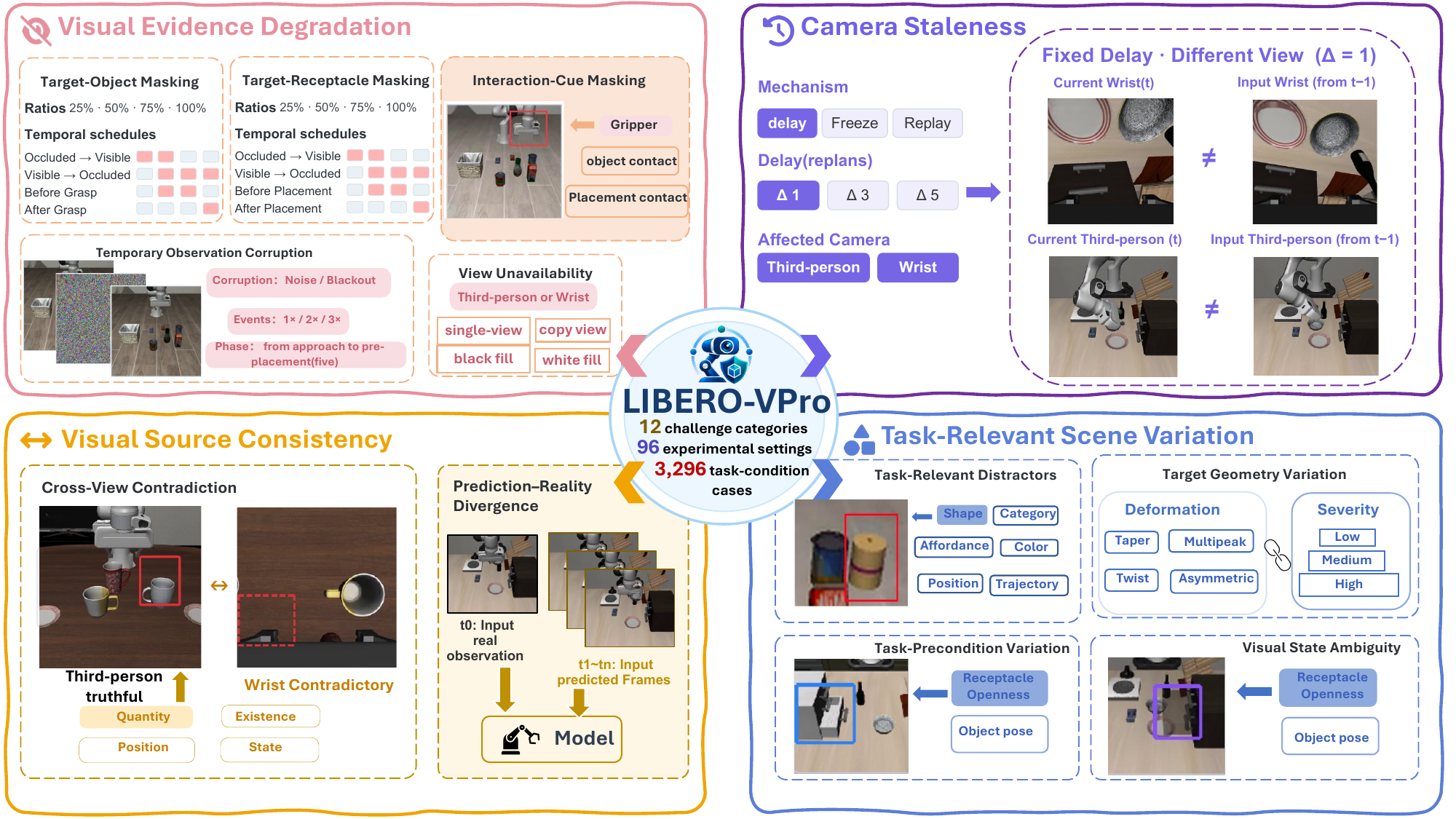}%
    \caption{Overview of LIBERO-VPro. The benchmark organizes 12 challenge categories into four complementary dimensions of closed-loop visual robustness: Visual Evidence Degradation, Camera Staleness, Visual Source Consistency, and Task-Relevant Scene Variation. LIBERO-VPro comprises 96 experimental settings and 3,296 task-condition cases.}%
    \label{fig:framework}%
  \end{minipage}%
}
\makeatother
\begin{document}
\bstctlcite{IEEEtranBSTCTL}
\maketitle
\thispagestyle{empty}
\pagestyle{empty}

\begin{abstract}

Robotic foundation models achieve impressive performance on standard manipulation benchmarks, yet these evaluations typically assume clean, timely, and consistent visual observations throughout execution. We introduce LIBERO-VPro, a benchmark for systematically evaluating the closed-loop visual robustness of robotic foundation models by perturbing the visual evidence available during execution. LIBERO-VPro covers four complementary dimensions, including Visual Evidence Degradation, Camera Staleness, Visual Source Consistency, and Task-Relevant Scene Variation, spanning 12 challenge categories, 96 experimental settings, and 3,296 task-condition cases. We evaluate three vision-language-action models and three world-action models over approximately 196,000 simulated episodes, complemented by 200 real-world rollouts on a Franka Research 3. Our results reveal that strong nominal performance can mask substantial weaknesses in visual grounding and adaptation. Models often remain successful despite severe object-level occlusion, yet degrade sharply when local interaction cues are disrupted or familiar spatial priors are violated. They are also highly sensitive to stale or missing observations and struggle when changed task preconditions require behavioral adaptation. Finally, VLAs and WAMs exhibit distinct robustness profiles, showing that visual robustness is multi-dimensional and architecture-dependent. LIBERO-VPro provides a systematic diagnostic framework for developing robotic foundation models that can more reliably ground and adapt their actions under challenging visual conditions.
\end{abstract}
\begin{table*}[t]
\caption{Comparison with representative robotic manipulation benchmarks in terms of visual robustness coverage. $\cmark$ indicates explicit coverage, $\pmark$ partial correspondence, and $\xmark$ no coverage.}
\label{tab:benchmark_comparison}
\centering
\scriptsize
\setlength{\tabcolsep}{2.1pt}
\renewcommand{\arraystretch}{1.12}
\resizebox{\textwidth}{!}{%
\begin{tabular}{l c c c c c c c c c c c c}
\toprule
\multirow{2}{*}{\raisebox{-2.4ex}{\textbf{Benchmark}}} &
\multicolumn{5}{c}{\textbf{Visual Evidence Degradation}} &
\multicolumn{2}{c}{\textbf{Visual Source Consistency}} &
\multicolumn{4}{c}{\textbf{Task-Relevant Scene Variation}} &
\multirow{2}{*}{\raisebox{-2.4ex}{\textbf{Camera Staleness}}} \\
\cmidrule(lr){2-6}\cmidrule(lr){7-8}\cmidrule(lr){9-12}
&
\colhdr{\textbf{Target-Obj.}}{\textbf{Mask.}} & \colhdr{\textbf{Target-Rec.}}{\textbf{Mask.}} &
\colhdr{\textbf{Interact.}}{\textbf{Cue Mask.}} & \colhdr{\textbf{Temp.}}{\textbf{Obs.\ Corr.}} &
\colhdr{\textbf{View}}{\textbf{Unavail.}} &
\colhdr{\textbf{Cross-View}}{\textbf{Contrad.}} &  
\colhdr{\textbf{Pred.--Reality}}{\textbf{Divergence}} &
\colhdr{\textbf{Task-Rel.}}{\textbf{Distr.}} & \colhdr{\textbf{Task Precond.}}{\textbf{Var.}} &
\colhdr{\textbf{Visual State}}{\textbf{Ambig.}} & \colhdr{\textbf{Target Geom.}}{\textbf{Var.}} & \\
\midrule
LIBERO~\cite{liu2023libero} &
\xmark & \xmark & \xmark & \xmark & \xmark & \xmark & \xmark & \xmark &
\xmark & \xmark & \xmark & \xmark \\
RoboTwin 2.0~\cite{chen2025robotwin2} &
\xmark & \xmark & \xmark & \xmark & \xmark & \xmark & \xmark & \pmark &
\xmark & \xmark & \xmark & \xmark \\
RoboCasa~\cite{nasiriany2024robocasa} &
\xmark & \xmark & \xmark & \xmark & \xmark & \xmark & \xmark & \xmark &
\xmark & \xmark & \xmark & \xmark \\
COLOSSEUM~\cite{pumacay2024colosseum} &
\xmark & \xmark & \xmark & \xmark & \xmark & \xmark & \xmark & \pmark &
\xmark & \xmark & \pmark & \xmark \\
VLATest~\cite{wang2025vlatest} &
\xmark & \xmark & \xmark & \xmark & \xmark & \xmark & \xmark & \pmark &
\xmark & \xmark & \xmark & \xmark \\
RobustVLA~\cite{chen2026robustvla} &
\xmark & \xmark & \xmark & \pmark & \xmark & \xmark & \xmark & \pmark &
\xmark & \xmark & \xmark & \xmark \\
LIBERO-PRO~\cite{zhou2025liberopro} &
\xmark & \xmark & \xmark & \xmark & \xmark & \xmark & \xmark & \xmark &
\pmark & \xmark & \pmark & \xmark \\
LIBERO-Plus~\cite{fei2026liberoplus} &
\xmark & \xmark & \xmark & \pmark & \xmark & \xmark & \xmark & \pmark &
\pmark & \xmark & \xmark & \xmark \\
VLA-Arena~\cite{VLA-Arena} &
\xmark & \xmark & \xmark & \pmark & \xmark & \xmark & \xmark & \pmark &
\xmark & \xmark & \xmark & \xmark \\
RoboTrustBench~\cite{li2026robotrustbench} &
\pmark & \pmark & \xmark & \xmark & \xmark & \xmark & \xmark & \pmark &
\pmark & \pmark & \xmark & \xmark \\
\midrule
\textbf{LIBERO-VPro (Ours)} &
\cmark & \cmark & \cmark & \cmark & \cmark & \cmark & \cmark & \cmark &
\cmark & \cmark & \cmark & \cmark \\
\bottomrule
\end{tabular}%
}
\end{table*}
\section{Introduction}
\label{sec:introduction}
Robotic foundation models are rapidly advancing toward general-purpose manipulation by integrating visual observations, language instructions, and action generation. Vision-language-action models (VLAs) connect visual and language representations to robot control~\cite{brohan2022rt1,brohan2023rt2,kim2024openvla, intelligence2025pi05}, while world-action models (WAMs) incorporate future world states prediction into policy learning~\cite{ye2026world, li2026lingbotva, yuan2026fastwam}. Benchmarks such as LIBERO~\cite{liu2023libero} and RoboTwin~\cite{chen2025robotwin2} have enabled standardized evaluation across diverse manipulation tasks. However, high task success under nominal conditions does not necessarily imply that a policy can robustly use visual feedback during closed-loop execution.

Standard manipulation benchmarks typically assume that observations remain complete, synchronized, and mutually consistent throughout a rollout. For physical deployment, nevertheless, the visual evidence available during execution may not always be reliable. Task-relevant objects can become occluded by the robot or surrounding clutter; camera streams can be temporarily corrupted, delayed, or unavailable; different viewpoints can provide conflicting evidence; and the physical scene can change in ways that invalidate previously successful action patterns. These failures occur inside the perception–action loop, where the policy must continuously interpret imperfect observations and adjust its behavior accordingly. Existing robustness benchmarks, such as LIBERO-Plus~\cite{fei2026liberoplus}, LIBERO-PRO~\cite{zhou2025liberopro} and VLA-Arena~\cite{VLA-Arena}, have substantially expanded evaluation beyond nominal settings by varying object appearance, lighting, viewpoints, initial states, instructions, and environments. Nevertheless, as shown in Table~\ref{tab:benchmark_comparison}, these evaluations primarily perturb the scene or task configuration, while the visual observations available to the policy during closed-loop execution generally remain temporally coherent and internally consistent. Consequently, an important question remains underexplored: \textit{How robust are robotic foundation models when the visual evidence they rely on becomes incomplete, stale, inconsistent, or behaviorally misleading during closed-loop execution?}

To address this question, we introduce \textbf{LIBERO-VPro}, a benchmark for closed-loop visual robustness of robotic foundation models built upon LIBERO. As shown in Fig.~\ref{fig:framework}, LIBERO-VPro consists of four complementary dimensions. These dimensions are designed to isolate four fundamental requirements for reliable closed-loop visual control, whether task-relevant evidence is available, temporally current, mutually consistent across sources, and sufficient to support behavioral adaptation when the scene changes. \textit{Visual Evidence Degradation} removes or corrupts task-relevant visual information at different spatial and temporal scales. \textit{Camera Staleness} introduces temporal mismatch between observations and the current physical state. \textit{Visual Source Consistency} introduces contradictory evidence across views or between predicted and realized observations. \textit{Task-Relevant Scene Variation} modifies visual properties that should require the policy to re-ground its behavior in the current scene. Together, these dimensions span 12 challenge categories, 96 experimental settings, and 3,296 task-condition cases, enabling controlled diagnosis of different sources of closed-loop visual failure.

We conduct a comprehensive evaluation using six representative robotic foundation models on LIBERO-VPro, including three VLAs, OpenVLA-OFT~\cite{kim2025openvlaoft}, $\pi_0$~\cite{black2024pi0}, $\pi_{0.5}$~\cite{intelligence2025pi05}, and three WAMs, FastWAM~\cite{yuan2026fastwam}, LingBot-VA~\cite{li2026lingbotva}, LaWAM~\cite{chen2026lawam}, over approximately 196,000 simulated episodes. The evaluation reveals several failure modes that are largely hidden by nominal task success. First, policies often remain successful when the target object or receptacle is occluded, but degrade sharply when interaction cues such as the gripper is removed. Second, models handle appearance-based distractors well but fail when distractors violate familiar spatial layouts, revealing strong reliance on learned spatial priors. Third, changing task preconditions is far more disruptive than introducing visual ambiguity, highlighting limited behavioral adaptation. Finally, VLAs and WAMs exhibit distinct robustness profiles, showing that visual robustness is inherently multi-dimensional. To examine whether these failures extend beyond simulation, we conduct 200 real-world rollouts on a Franka Research 3 using $\pi_{0.5}$ and LaWAM across two manipulation tasks. Representative perturbations from all four LIBERO-VPro dimensions reveal broadly consistent trends. Both models are highly sensitive to missing views, cross-view inconsistencies, and violations of familiar spatial layouts, while robustness to stale observations varies by task and model. These results indicate that LIBERO-VPro captures failure modes that also arise in physical closed-loop manipulation.

Our main contributions are as follows:
\begin{itemize}
\item We introduce LIBERO-VPro, a benchmark that systematically perturbs visual observations \emph{during} closed-loop execution along four complementary dimensions, covering 12 challenge categories, 96 experimental settings, and 3{,}296 task-condition cases.
\item We conduct a comprehensive evaluation of three VLA and three WAM models, producing approximately 196{,}000 simulated episodes, complemented by 200 physical rollouts on a Franka Research 3, providing controlled analysis of visual robustness across both simulation and real-world manipulation.
\item We reveal strong reliance on spatial priors, vulnerability to disrupted interaction cues and scene changes, and distinct robustness profiles across VLAs and WAMs.
\end{itemize}

\section{Related Work}
\label{sec:related_work}
\subsection{Robotic Foundation Models}
  Vision-language-action (VLA) models extend vision-language pretraining to continuous robot control, from RT-1/
  RT-2~\cite{brohan2022rt1,brohan2023rt2} to OpenVLA~\cite{kim2024openvla} and embodied multimodal
  reasoning~\cite{driess2023palme}. Our VLA baselines include OpenVLA-OFT~\cite{kim2025openvlaoft}, $
  \pi_0$~\cite{black2024pi0}, and $\pi_{0.5}$~\cite{intelligence2025pi05}. World-action models (WAMs) augment
  control with future-state prediction or latent dynamics: GR-2~\cite{cheang2024gr2},
  FastWAM~\cite{yuan2026fastwam}, LingBot-VA~\cite{li2026lingbotva}, and LaWAM~\cite{chen2026lawam} represent
  video-generation pretraining, test-time imagination, causal world modeling, and latent-dynamics policy
  generation, respectively. VLAs map observations directly to actions, whereas WAMs explicitly model future
  dynamics, motivating their joint robustness evaluation.
  \subsection{Robotic Manipulation Benchmarks}
  LIBERO~\cite{liu2023libero} and nominal
  simulators~\cite{james2019rlbench,gu2023maniskill2,zhu2020robosuite,yu2020metaworld,mees2022calvin} provide
  standardized manipulation tasks, while MimicGen~\cite{mandlekar2023mimicgen}, GenAug~\cite{chen2023genaug},
  and RoboCasa~\cite{nasiriany2024robocasa} scale data and scene diversity. Recent benchmarks evaluate
  appearance, lighting, background, and viewpoint changes~\cite{pumacay2024colosseum}, progressive
  difficulty~\cite{fei2026liberoplus}, object, initial-state, instruction, and environment
  shifts~\cite{zhou2025liberopro}, domain randomization~\cite{chen2025robotwin2,tobin2017domainrandom},
  confounding objects~\cite{wang2025vlatest}, multimodal perturbations~\cite{chen2026robustvla}, or offline
  generated-video quality~\cite{li2026robotrustbench}. Existing methods primarily modify scene, task, or generic
  visual conditions without testing our stage-specific transient corruption, cross-view contradiction, or
  recursive predicted-feedback settings. LIBERO-VPro instead perturbs visual availability, temporal freshness,
  cross-source consistency, and task-relevant scene content during execution.
  Table~\ref{tab:benchmark_comparison} compares their coverage.

\section{Benchmark Construction}
\label{sec:Benchmark}
\begin{figure}[t]
  \centering
  \includegraphics[width=0.9\columnwidth]{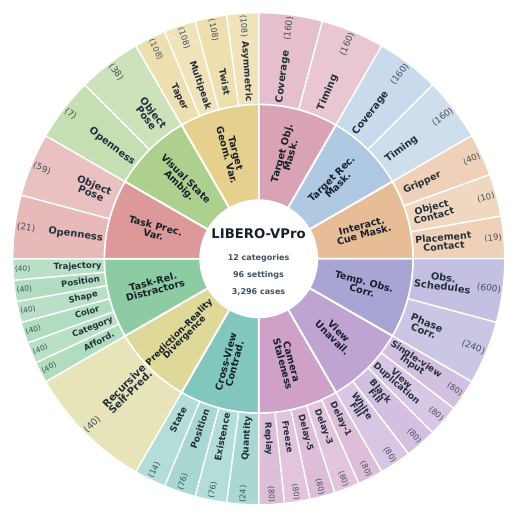}
  \caption{LIBERO-VPro benchmark hierarchy: 12 challenge categories decomposed into 96 experimental settings totaling 3,296 cases.}
  \label{fig:hierarchy}
\end{figure}

LIBERO-VPro evaluates closed-loop visual robustness by systematically modifying the visual evidence available to a manipulation policy during execution. As shown in Fig.~\ref{fig:framework}, LIBERO-VPro organizes visual challenges into four complementary dimensions: Visual Evidence Degradation, Camera Staleness, Visual Source Consistency, and Task-Relevant Scene Variation. The benchmark spans 12 challenge categories, 96 experimental settings, and 3,296 task-condition cases. Fig.~\ref{fig:hierarchy} summarizes the complete hierarchy.

\subsection{Visual Evidence Degradation} 
Reliable manipulation should not depend on every visual cue remaining continuously available. In realistic execution, task-relevant entities may be occluded by the robot, manipulated objects, or surrounding clutter, while camera observations may be temporarily corrupted or lost. We therefore evaluate information loss at multiple spatial scales, from individual objects and interaction regions to complete frames and camera views.

\subsubsection{Target-Object and Target-Receptacle Masking}
We separately occlude the target object and target receptacle to test whether policies depend on their online visual appearance during execution. Each category includes four masking ratios (25\%, 50\%, 75\% and 100\%) and four temporal schedules that control when the relevant entity becomes visible or occluded, including transitions around grasping or placement.

\subsubsection{Interaction-Cue Masking}
We selectively remove local interaction cues while preserving the surrounding scene. Perturbations target the robot gripper, gripper-object contact region, and placement contact region, enabling us to isolate the importance of fine-grained visual feedback during manipulation.

\subsubsection{Temporary Observation Corruption}
We introduce transient full-frame corruption using either blackout or visual noise. Corruption is controlled by both occurrence frequency and manipulation phase from approach to pre-placement.

\subsubsection{View Unavailability}
We selectively remove either the third-person or wrist-camera observation. We consider single-view input as well as replacements using constant images or duplicated observations from the remaining camera.

\subsection{Camera Staleness}
Closed-loop control requires observations to reflect the robot's current physical state. In practice, camera latency, frame drops, and asynchronous sensor pipelines can cause policies to act on outdated observations, potentially producing actions that are individually plausible but inappropriate for the present state. We simulate such temporal mismatch through three fixed-delay levels, observation freezing, and historical replay. Each perturbation is independently applied to either the third-person or wrist camera while the other view remains current. This design allows us to measure both overall sensitivity to stale observations and the relative dependence of each policy on individual camera streams.

\subsection{Visual Source Consistency}
Robotic foundation models increasingly integrate information from multiple visual sources, including external camera views (third-person and wrist) and, for WAMs, internally predicted future observations. Reliable control requires these sources to remain consistent about the task-relevant state of the environment. Unlike Camera Staleness, which tests whether an observation is temporally current, Visual Source Consistency examines failures caused by disagreement between different sources of visual evidence. We consider two settings: conflicts across physical camera views and divergence between predicted and realized visual futures.

\subsubsection{Cross-View Contradiction}

Multi-view observations may provide inconsistent evidence due to occlusion, viewpoint-dependent visibility, or perception errors. We introduce controlled contradictions between the third-person and wrist cameras along four task-relevant attributes: quantity, existence, position, and state. One view is manipulated while the other remains truthful, with both kept temporally aligned. This setting tests whether policies can reconcile conflicting visual evidence rather than simply rely on a preferred camera view.

\subsubsection{Prediction–Reality Divergence}
WAMs may additionally rely on internally predicted future observations for action generation, making the consistency between imagined and realized scene evolution critical for closed-loop control. We therefore provide a ground-truth observation only at initialization and recursively feed the model's predicted frames back as subsequent visual input. This setting examines whether prediction errors accumulate into control failures when imagined futures increasingly replace observations of the realized environment. Because explicit future-frame generation is required, this evaluation is applied only to compatible WAMs.

\subsection{Task-Relevant Scene Variation}
The previous three dimensions perturb the quality, timeliness, or consistency of visual observations while largely preserving the underlying task configuration. In contrast, Task-Relevant Scene Variation changes the scene itself while keeping observations complete, current, and mutually consistent. It therefore tests whether a policy can re-ground its behavior in the current scene rather than rely on visual or action patterns memorized from the training distribution.

\subsubsection{Task-Relevant Distractors}
A robust manipulation policy should identify the instructed target from current visual evidence rather than rely on superficial similarity or familiar spatial layouts. We introduce distractors that share the target's category, color, shape, or affordance, place a competing object at the target's original position, or obstruct its typical manipulation trajectory respectively. These conditions probe robustness to appearance-based confusion as well as reliance on learned spatial and motion priors.

\subsubsection{Task-Precondition Variation}
Manipulation strategies often depend on the current state of the environment. When that state changes, successful execution may require additional or altered actions. We vary two task-relevant preconditions, including receptacle openness and object pose, while keeping the language instruction unchanged. For example, placing an object into a closed receptacle requires first opening it. This scenario tests whether policies adapt their action sequence to the observed scene rather than replay a nominal manipulation routine.

\subsubsection{Visual State Ambiguity}
A related but distinct challenge arises when the physical state is unchanged but its visual evidence becomes difficult to interpret. We therefore preserve the underlying task state while weakening the cues associated with openness and object pose, so that multiple interpretations become visually plausible.  We apply the same two attributes as Task-Precondition Variation but render multiple states simultaneously in the observation, such as a bowl appearing both upright and inverted.

\subsubsection{Target Geometry Variation}
Objects within the same semantic category can exhibit substantial geometric variation in the real world, requiring manipulation policies to adjust grasp locations and end-effector poses accordingly. We introduce four target-object deformations, including Taper, Multipeak, Twist, and Asymmetric, each at three severity levels while preserving object category. This scenario evaluates whether policies adapt manipulation behavior to the observed geometry rather than reuse shape-specific grasp strategies learned during training.

\newcolumntype{G}{!{\color{black}\vrule width 0.3pt}}
\newcolumntype{H}{!{\color{black}\vrule width \arrayrulewidth}}
\begin{table*}[t]
  \caption{Success rate (SR, \%) across the eleven LIBERO-VPro challenge categories shared by all six models. Normal denotes clean-condition performance, while gray $\downarrow$ values report the absolute SR drop relative to Normal. Categories marked $\uparrow$ are robustness indicators, where higher SR indicates greater resilience. Categories marked $\diamond$ are diagnostic indicators, where high SR may reflect reliance on non-visual cues or learned priors rather than genuine visual grounding.}
  \label{tab:category_matrix}
  \centering
  \footnotesize
  \setlength{\tabcolsep}{1.5pt}
  \renewcommand{\arraystretch}{1.02}
  \newcommand{\cathead}[2]{%
    \shortstack[c]{\strut #1\\\strut #2\strut}}
  \resizebox{\textwidth}{!}{%
  \begin{tabular}{clGcG*{5}{c}GcG*{4}{c}GcGc}
    \hline
    & \multirow{2}{*}{Model}
      & \multirow{2}{*}{Normal}
      & \multicolumn{5}{cG}{\raisebox{4pt}{Visual Evidence Degradation}}
      & \multicolumn{1}{cG}{\shortstack[c]{Visual Source\\Consistency}}
      & \multicolumn{4}{cG}{\raisebox{4pt}{Task-Relevant Scene Variation}}
      & \multirow{2}{*}{\shortstack[c]{Camera$\,\uparrow$\\Staleness}}
      & \multirow{2}{*}{\shortstack[c]{Overall\\Avg.}} \\
    \noalign{\vskip-2pt}
    \cline{4-8}\cline{9-9}\cline{10-13}
    & &
      & \cathead{Target Obj.$\,\diamond$}{Mask.}
      & \cathead{Target Rec.$\,\diamond$}{Mask.}
      & \cathead{Interact.$\,\diamond$}{Cue Mask.}
      & \cathead{Temp. Obs.$\,\uparrow$}{Corr.}
      & \cathead{View$\,\uparrow$}{Unavail.}
      & \cathead{Cross-View$\,\diamond$}{Contrad.}
      & \cathead{Task-Rel.$\,\uparrow$}{Distractors}
      & \cathead{Task Prec.$\,\uparrow$}{Var.}
      & \cathead{Visual State$\,\diamond$}{Ambig.}
      & \cathead{Target$\,\uparrow$}{Geom. Var.}
      & & \\
    \hline
    \multirow{3}{*}{\scriptsize VLA}
    & OpenVLA-OFT  & 97.0 & 91.28\drop{5.7} & 54.34\drop{42.7} & 16.23\drop{80.8} & 76.39\drop{20.6} & \best{29.25}\drop{67.8} & 54.47\drop{42.5} & 65.08\drop{31.9} & 39.38\drop{57.6} & 88.22\drop{8.8} & 81.74\drop{15.3} & 24.02\drop{73.0} & 56.40\drop{40.6} \\
    & $\pi_0$      & 92.5 & 81.00\drop{11.5} & 87.50\drop{5.0} & \best{71.59}\drop{20.9} & 70.94\drop{21.6} & \second{28.19}\drop{64.3} & 51.53\drop{41.0} & 61.25\drop{31.2} & 37.13\drop{55.4} & 70.67\drop{21.8} & 79.31\drop{13.2} & \second{27.00}\drop{65.5} & 60.55\drop{31.9} \\
    & $\pi_{0.5}$  & 97.2 & \second{93.41}\drop{3.8} & \second{95.66}\drop{1.5} & 47.54\drop{49.7} & \second{82.80}\drop{14.4} & 22.72\drop{74.5} & 59.21\drop{38.0} & \best{70.92}\drop{26.3} & \best{45.50}\drop{51.7} & 85.78\drop{11.4} & 85.35\drop{11.9} & 20.52\drop{76.7} & 64.49\drop{32.7} \\
    \hline
    \multirow{3}{*}{\scriptsize WAM}
    & FastWAM      & 98.8 & 93.38\drop{5.4} & 95.09\drop{3.7} & 69.42\drop{29.4} & 81.34\drop{17.5} & 10.12\drop{88.7} & \second{63.63}\drop{35.2} & 69.50\drop{29.3} & \second{43.63}\drop{55.2} & \best{94.67}\drop{4.1} & \best{88.24}\drop{10.6} & 17.00\drop{81.8} & \second{66.00}\drop{32.8} \\
    & LingBot-VA   & 97.0 & 84.69\drop{12.3} & \best{96.78}\drop{0.2} & \second{70.43}\drop{26.6} & 67.89\drop{29.1} & 21.00\drop{76.0} & \best{70.11}\drop{26.9} & \second{69.75}\drop{27.2} & \best{45.50}\drop{51.5} & \second{93.33}\drop{3.7} & 86.39\drop{10.6} & \best{34.50}\drop{62.5} & \best{67.31}\drop{29.7} \\
    & LaWAM        & 98.5 & \best{93.44}\drop{5.1} & 91.06\drop{7.4} & 69.13\drop{29.4} & \best{85.97}\drop{12.5} & 17.97\drop{80.5} & 58.37\drop{40.1} & 69.50\drop{29.0} & 41.38\drop{57.1} & 91.33\drop{7.2} & \second{87.75}\drop{10.8} & 16.60\drop{81.9} & 65.68\drop{32.8} \\
    \hline
  \end{tabular}%
  }
\end{table*}
\section{Experiments}
\label{sec:experiments}
\subsection{Experimental Setup}
\label{sec:experimental_setup}

\textit{Evaluated Models.}
We evaluate six representative robotic foundation models: three VLAs, including OpenVLA-OFT~\cite{kim2025openvlaoft}, $\pi_0$~\cite{black2024pi0}, and $\pi_{0.5}$~\cite{intelligence2025pi05}, as well as three WAMs, including FastWAM~\cite{yuan2026fastwam}, LingBot-VA~\cite{li2026lingbotva}, and LaWAM~\cite{chen2026lawam}. All models are trained on the standard LIBERO demonstrations~\cite{liu2023libero}. We use the officially released checkpoints and inference pipelines whenever available. Since LingBot-VA only releases a LIBERO-Long checkpoint, we additionally train it on LIBERO-Object, LIBERO-Spatial, and LIBERO-Goal using the official training recipe and default hyperparameters.

\textit{Evaluation Protocol.}
Experiments cover 40 manipulation tasks from four LIBERO suites: LIBERO-Long, LIBERO-Object, LIBERO-Spatial, and LIBERO-Goal. The 12 challenge categories in LIBERO-VPro yield 96 experimental settings and 3,296 task-condition cases. Each case is evaluated over ten trials with different initial configurations, resulting in approximately 196,000 episodes in total. Unless otherwise specified, perturbations affect only the visual observations.

\textit{Metric and Interpretation.} We report task Success Rate (SR) to evaluate the performance. For conventional robustness conditions, such as camera staleness or view unavailability, higher SR indicates stronger resilience. However, several scenarios, including object masking, interaction-cue masking, cross-view contradiction, and visual state ambiguity, are primarily diagnostic. High success may indicate robustness, but can also arise from reliance on memorized spatial priors, alternative camera views, or non-visual cues. We therefore interpret these conditions together with their controlled comparisons rather than treating all SR values as uniformly higher-is-better.

\subsection{Overall Results}
\label{sec:overall}

Table~\ref{tab:category_matrix} summarizes performance across the 11 challenge categories shared by all six models. First, visual robustness is highly scenario-dependent. No model consistently dominates across all categories, showing that robustness is multi-dimensional rather than a single capability that scales uniformly with model performance. For example, $\pi_{0.5}$ performs strongly under semantic distractors and task precondition variation but is highly vulnerable to camera staleness and missing views. Second, all models are surprisingly tolerant to removal of object-level evidence like target-object and target-receptacle masking, which often preserve high success rate. This tolerance likely reflects reliance on learned spatial priors and remaining scene cues. In contrast, masking local interaction cues causes substantially larger degradation, indicating that fine manipulation depends more strongly on visual feedback around the robot-object interaction. Third, View Unavailability and Camera Staleness are among the most disruptive conditions, showing that both the availability and temporal freshness of visual observations are critical for closed-loop control. Notably, all three VLAs outperform all three WAMs under View Unavailability, suggesting that the evaluated WAMs are particularly sensitive to changes in their expected multi-view input structure.

\subsection{What Visual Evidence Do Policies Rely On?}
\label{sec:visual_evidence}

\begin{figure}[t]
\centering
\includegraphics[width=.98\columnwidth]{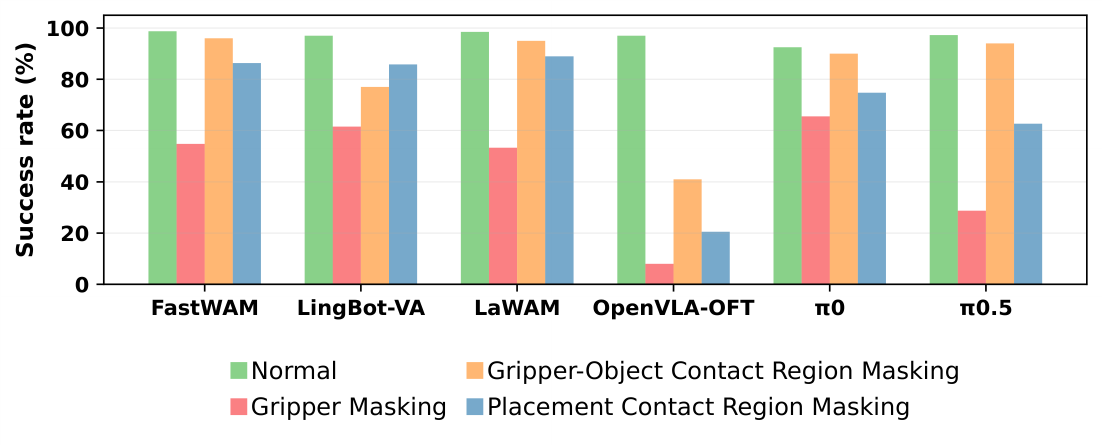}
\caption{Success rate (\%) under Interaction-Cue Masking.}
\label{fig:extended_interaction}
\end{figure}

\textit{Interaction Cues Matter More Than Static Object Appearance.} We first examine how performance changes when visual evidence is removed at different spatial scales. Target-object and target-receptacle masking preserve relatively high success for most models, even under severe occlusion. In contrast, Interaction-Cue Masking causes substantially larger degradation. Fig.~\ref{fig:extended_interaction} further decomposes the interaction region. Across all six models, masking the gripper is the most disruptive intervention, whereas masking the target object or receptacle is considerably less harmful. This contrast indicates that policies can often compensate for missing object evidence using learned task and spatial priors, but remain strongly dependent on local visuomotor feedback around the end effector.

\begin{figure}[t]
\centering
\includegraphics[width=0.98\columnwidth]{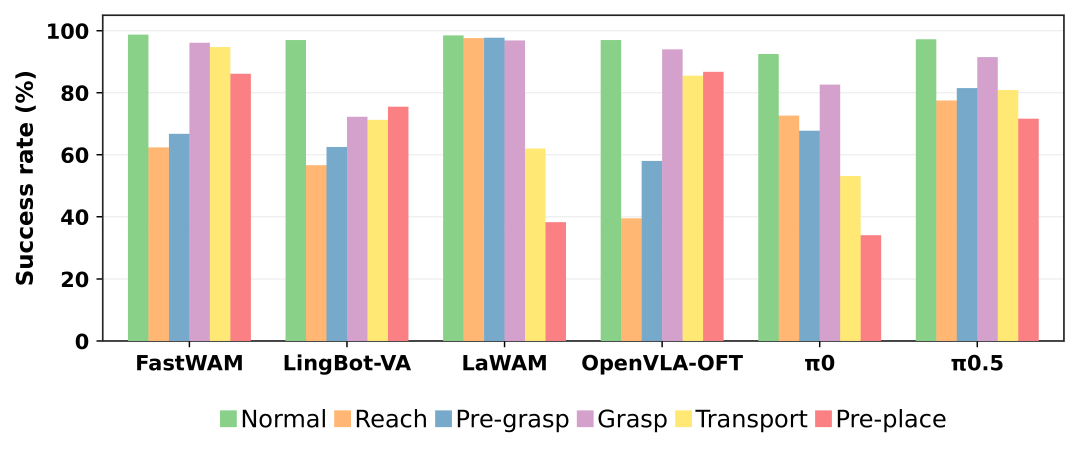}
\caption{Success rate (\%) under phase-triggered Observation Corruption.}
\label{fig:extended_phase}
\end{figure}

\textit{Sensitivity to Visual Loss Depends on Execution Phase.} Transient corruption provides a complementary temporal view of visual dependence. Fig.~\ref{fig:extended_phase} shows that corruption during the Grasp phase has relatively limited impact, whereas corruption during reaching, transport, and pre-placement produces larger and strongly model-dependent degradation. The temporal profiles also differ across policies. OpenVLA-OFT is more sensitive to early-stage corruption, while $\pi_0$ and LaWAM degrade more strongly after grasping. These results show that visual robustness depends not only on how much information is lost, but also on when it becomes unavailable.

\subsection{How Do Policies Use Multi-View Visual Evidence?}
\label{sec:camera_roles}
\begin{table}[t]
  \caption{Per-view success rate (\%) under Camera Staleness and View Unavailability. For Staleness, only the indicated camera stream is made stale while the other remains current. For Availability, only the indicated view is retained and the other is removed.}
  \label{tab:view_reliance}
  \centering
  \scriptsize
  \setlength{\tabcolsep}{1.5pt}
  \resizebox{0.92\columnwidth}{!}{%
  \begin{tabular}{@{}cl|c|@{}p{2pt}@{}*{2}{c}@{}p{2pt}@{}p{2pt}@{}p{2pt}@{}*{2}{c}@{}p{2pt}@{}}
    \hline
    & \multirow{2}{*}{Model} & \multirow{2}{*}{Normal} & & \multicolumn{2}{c}{Staleness} & & & & \multicolumn{2}{c}{Availability} & \\
    \hhline{~~|~|~--~~~--~}
    & & & & \multicolumn{1}{c}{Third} & \multicolumn{1}{c}{Wrist}
      & & & & \multicolumn{1}{c}{Third} & \multicolumn{1}{c}{Wrist} & \\
    \hline
    \multirow{3}{*}{\scriptsize VLA}
    & OpenVLA-OFT & 97.0 & & \second{37.15}\drop{59.9} & 10.90\drop{86.1} & & & & \second{15.31}\drop{81.7} & 43.19\drop{53.8} & \\
    & $\pi_0$ & 92.5 & & 32.85\drop{59.6} & \second{21.15}\drop{71.3} & & & & 7.62\drop{84.9} & \best{48.75}\drop{43.8} & \\
    & $\pi_{0.5}$ & 97.2 & & 21.35\drop{75.8} & 19.70\drop{77.5} & & & & 2.19\drop{95.0} & \second{43.25}\drop{54.0} & \\
    \hline
    \multirow{3}{*}{\scriptsize WAM}
    & FastWAM & 98.8 & & 21.65\drop{77.2} & 12.35\drop{86.5} & & & & 0.56\drop{98.2} & 19.69\drop{79.1} & \\
    & LingBot-VA & 97.0 & & \best{55.75}\drop{41.2} & 13.25\drop{83.8} & & & & 1.38\drop{95.6} & 40.62\drop{56.4} & \\
    & LaWAM & 98.5 & & 9.95\drop{88.5} & \best{23.25}\drop{75.2} & & & & \best{35.00}\drop{63.5} & 0.94\drop{97.6} & \\
    \hline
  \end{tabular}%
  }
\end{table}

\textit{Stale and Missing Views Reveal Strong Camera Dependence.} 
Table~\ref{tab:view_reliance} reveals substantial asymmetry in how policies use the third-person and wrist cameras. Five of the six models are more sensitive to perturbations of the wrist view, whereas LaWAM shows stronger dependence on the third-person camera. Camera Staleness and View Unavailability expose complementary vulnerabilities. Staleness provides visually valid but outdated observations, whereas view loss removes a source entirely. Strong degradation under both conditions indicates that models rely heavily on the expected temporal and multi-view structure of their inputs rather than flexibly reallocating attention to the remaining evidence.

\begin{table}[t]
\centering
\scriptsize
\caption{Success rate (\%) under Cross-View Contradiction.}
\label{tab:extended_contradiction}
\setlength{\tabcolsep}{2pt}
\resizebox{0.92\columnwidth}{!}{%
\begin{tabular}{@{}clHcH*{4}{c}@{}}
\hline
& Model & Normal & Quantity & Existence & Position & State \\
\hline
\multirow{3}{*}{\scriptsize VLA}
& OpenVLA-OFT & 97.0 & 76.7\drop{20.3} & 53.3\drop{43.7} & 46.5\drop{50.5} & 66.4\drop{30.6} \\
& $\pi_0$ & 92.5 & 69.6\drop{22.9} & 47.1\drop{45.4} & 47.2\drop{45.3} & 67.9\drop{24.6} \\
& $\pi_{0.5}$ & 97.2 & 79.2\drop{18.0} & 56.8\drop{40.4} & 52.5\drop{44.7} & \second{74.3}\drop{22.9} \\
\hline
\multirow{3}{*}{\scriptsize WAM}
& FastWAM & 98.8 & 79.6\drop{19.2} & 59.6\drop{39.2} & \second{63.0}\drop{35.8} & 61.4\drop{37.4} \\
& LingBot-VA & 97.0 & \best{85.0}\drop{12.0} & \best{68.8}\drop{28.2} & \best{65.8}\drop{31.2} & \best{75.0}\drop{22.0} \\
& LaWAM & 98.5 & \second{81.7}\drop{16.8} & \second{61.3}\drop{37.2} & 46.3\drop{52.2} & 67.9\drop{30.6} \\
\hline
\end{tabular}%
}
\end{table}

\textit{Conflicting Views Are Hardest When They Disagree Spatially.} We next examine Cross-View Contradiction, where temporally aligned views provide incompatible evidence about object quantity, existence, position, or state. Table~\ref{tab:extended_contradiction} shows that quantity contradiction is generally the least disruptive, while existence and position contradictions cause substantially larger degradation. Importantly, high success under contradiction does not necessarily imply successful conflict resolution. Quantity or state inconsistencies may preserve the target location, allowing policies to execute the task without explicitly resolving the disagreement. By contrast, position contradiction provides competing positive evidence at different locations, directly interfering with the spatial representation needed for action generation.

\begin{figure}[t]
  \centering
  \includegraphics[width=0.9\columnwidth]{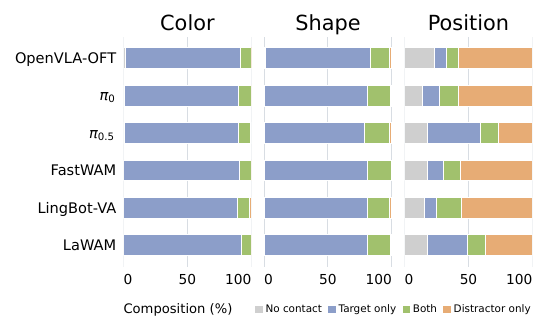}
  \caption{Robot-object contact composition under three Task-Relevant distractors. Each bar partitions trials by whether the gripper contacts the target, the distractor, both, or neither.}
  \label{fig:a4_contact_states}
\end{figure}

\subsection{Can Policies Re-Ground When the Scene Changes?}
\label{sec:scene_variation}
\begin{table}[t]
  \caption{Success rate (\%) under Task-Relevant Distractors. Each column corresponds to a different distractor type targeting the target object's category, color, shape, affordance, spatial position, or manipulation trajectory.}
  \label{tab:distractor_subclasses}
  \centering
  \scriptsize
  \setlength{\tabcolsep}{2pt}
  \resizebox{\columnwidth}{!}{%
  \begin{tabular}{@{}clHcH*{6}{c}@{}}
    \hline
    & Model & Normal & Category & Color & Shape & Afford. & Position & Trajectory \\
    \hline
    \multirow{3}{*}{\scriptsize VLA}
    & OpenVLA-OFT & 97.0 & 80.00\drop{17.0} & 92.75\drop{4.2} & 88.50\drop{8.5} & \second{93.25}\drop{3.8} & 4.25\drop{92.8} & 31.75\drop{65.2} \\
    & $\pi_0$ & 92.5 & 80.25\drop{12.2} & 86.25\drop{6.2} & 84.25\drop{8.2} & 89.00\drop{3.5} & 2.50\drop{90.0} & 25.25\drop{67.2} \\
    & $\pi_{0.5}$ & 97.2 & 88.25\drop{9.0} & 93.00\drop{4.2} & 91.75\drop{5.5} & \best{94.25}\drop{3.0} & \best{22.25}\drop{75.0} & 36.00\drop{61.2} \\
    \hline
    \multirow{3}{*}{\scriptsize WAM}
    & FastWAM & 98.8 & \best{92.25}\drop{6.5} & \best{94.50}\drop{4.3} & \best{92.75}\drop{6.0} & \best{94.25}\drop{4.5} & 4.75\drop{94.0} & \second{38.50}\drop{60.3} \\
    & LingBot-VA & 97.0 & \best{92.25}\drop{4.8} & 91.50\drop{5.5} & \second{92.50}\drop{4.5} & 86.75\drop{10.2} & 8.00\drop{89.0} & \best{47.50}\drop{49.5} \\
    & LaWAM & 98.5 & \second{88.75}\drop{9.8} & \second{94.00}\drop{4.5} & 90.75\drop{7.8} & \best{94.25}\drop{4.2} & \second{14.50}\drop{84.0} & 34.75\drop{63.8} \\
    \hline
  \end{tabular}%
  }
\end{table}

\textit{Spatial Distractors Expose Reliance on Learned Positional Priors.}
Table~\ref{tab:distractor_subclasses} lists the results of six distractors in the scene. Distractors sharing the target's category, color, shape, or affordance cause relatively limited degradation. In contrast, placing a distractor at the target's familiar position causes success to collapse across models, while trajectory obstruction causes the second-largest degradation. This pattern indicates that semantic similarity alone is not the dominant challenge. Instead, policies rely strongly on learned spatial priors. When a competitor occupies the target's familiar location, models frequently act toward the memorized position rather than re-grounding the target from the current observation.  Fig.~\ref{fig:a4_contact_states} shows the corresponding behavioral shift, with position displacement substantially increasing distractor-only contact, whereas semantic distractors (e.g., color and shape) more often preserve target-directed interaction.

\begin{table}[t]
\caption{Success rate (\%) comparing Task-Precondition Variation
  with Visual State Ambiguity over the shared attributes of receptacle openness and object pose.}
\label{tab:extended_precondition}
\centering
\scriptsize
\setlength{\tabcolsep}{3pt}
\resizebox{\columnwidth}{!}{%
\begin{tabular}{@{}cl|c|@{}p{2pt}@{}*{2}{c}@{}p{2pt}@{}p{2pt}@{}p{2pt}@{}*{2}{c}@{}p{2pt}@{}}
\hline
& \multirow{2}{*}{Model} & \multirow{2}{*}{Normal} & & \multicolumn{2}{c}{\shortstack{Task-Precondition\\Variation}} & & & & \multicolumn{2}{c}{\shortstack{Visual State\\Ambiguity}} & \\
\hhline{~~|~|~--~~~--~}
& & & & \multicolumn{1}{c}{Openness} & \multicolumn{1}{c}{Object Pose} & & & & \multicolumn{1}{c}{Openness} & \multicolumn{1}{c}{Object Pose} & \\
\hline
\multirow{3}{*}{\scriptsize VLA}
& OpenVLA-OFT & 97.0 & & 58.6\drop{38.4} & 25.3\drop{71.7} & & & & 95.7\drop{1.3} & 86.8\drop{10.2} & \\
& $\pi_0$ & 92.5 & & 51.9\drop{40.6} & 24.7\drop{67.8} & & & & 81.4\drop{11.1} & 68.7\drop{23.8} & \\
& $\pi_{0.5}$ & 97.2 & & \second{64.8}\drop{32.4} & \best{30.3}\drop{66.9} & & & & \second{97.1}\drop{0.1} & 83.7\drop{13.5} & \\
\hline
\multirow{3}{*}{\scriptsize WAM}
& FastWAM & 98.8 & & 59.5\drop{39.3} & \second{29.7}\drop{69.1} & & & & \best{100.0}\nodrop & \best{93.7}\drop{5.1} & \\
& LingBot-VA & 97.0 & & \best{67.6}\drop{29.4} & 29.3\drop{67.7} & & & & 94.3\drop{2.7} & \second{93.2}\drop{3.8} & \\
& LaWAM & 98.5 & & 61.9\drop{36.6} & 26.6\drop{71.9} & & & & \second{97.1}\drop{1.4} & 90.3\drop{8.2} & \\
\hline
\end{tabular}%
}
\end{table}

\textit{Task-Precondition Variation Is Harder Than Visual Ambiguity.} Task-Precondition Variation and Visual State Ambiguity provide a controlled comparison over the same two attributes: receptacle openness and object pose. The former changes the physical scene, whereas the latter preserves the underlying state but weakens its visual evidence. Table~\ref{tab:extended_precondition} shows that actual precondition changes cause severe degradation, while visual ambiguity preserves substantially higher performance for both openness and object pose. The results suggest that the primary difficulty is not simply uncertainty in visual interpretation, but adapting the manipulation strategy when the scene requires a different action sequence. For example, placing an object into a closed receptacle requires an additional opening action before the nominal placement behavior. Current policies often fail to make such scene-conditioned adjustments, revealing limited behavioral adaptation beyond familiar training trajectories.

\begin{table}[t]
\caption{Success rate (\%) under Target Geometry Variation. }
\label{tab:shape_variation}
\centering
\scriptsize
\setlength{\tabcolsep}{3pt}
\resizebox{\columnwidth}{!}{%
\begin{tabular}{@{}clHcH*{4}{c}Hc@{}}
\hline
& Model & Normal & Taper & Multipeak & Twist & Asymmetric & Overall Avg. \\
\hline
\multirow{3}{*}{\scriptsize VLA}
& OpenVLA-OFT & 97.2 & 80.6\drop{16.6} & 79.6\drop{17.6} & 75.9\drop{21.3} & 85.2\drop{12.0} & 80.3\drop{16.9} \\
& $\pi_0$ & 93.1 & 83.3\drop{9.8} & 84.3\drop{8.8} & 70.4\drop{22.7} & 81.5\drop{11.6} & 79.9\drop{13.2} \\
& $\pi_{0.5}$ & 97.5 & \second{90.7}\drop{6.8} & 88.0\drop{9.5} & 79.6\drop{17.9} & 88.0\drop{9.5} & \second{86.6}\drop{10.9} \\
\hline
\multirow{3}{*}{\scriptsize WAM}
& FastWAM & 98.6 & \best{89.5}\drop{9.1} & \best{89.7}\drop{8.9} & 82.1\drop{16.5} & \best{91.6}\drop{7.0} & \best{88.2}\drop{10.4} \\
& LingBot-VA & 96.7 & 86.5\drop{10.2} & 85.9\drop{10.8} & \best{85.5}\drop{11.2} & 87.7\drop{9.0} & 86.4\drop{10.3} \\
& LaWAM & 98.6 & \second{89.5}\drop{9.1} & \second{89.3}\drop{9.3} & \second{82.9}\drop{15.7} & \second{89.4}\drop{9.2} & 87.8\drop{10.8} \\
\hline
\end{tabular}%
}
\end{table}

\textit{Geometry Variation Is Comparatively Well Tolerated.} Table~\ref{tab:shape_variation} evaluates four target-geometry transformations. Twist consistently produces the largest degradation, while asymmetric deformation is generally the least disruptive. Nevertheless, performance remains relatively strong compared with positional distractors and task-precondition changes. The results show that current robotic foundation models generalize reasonably well to moderate geometric variation, while remaining more vulnerable to changes that invalidate learned spatial priors.

\subsection{Can a WAM Rely on Its Own Predicted Futures?}
\label{sec:imagined_future}
The previous experiments evaluate challenges shared by VLAs and WAMs. We finally investigate a capability specific to WAMs that explicitly generate future observations: whether their predicted futures remain sufficiently grounded to support closed-loop control. Starting from a real observation, we recursively replace subsequent visual inputs with the model's own predictions. Among the evaluated models, this experiment applies to LingBot-VA only.

\begin{table}[t]
\caption{Suite-wise success rate (\%) of LingBot-VA under normal observation and recursive predicted-frame feedback.}
\label{tab:recursive_suite}
\centering
\scriptsize
\setlength{\tabcolsep}{3pt}
\begin{tabular}{@{}lH*{4}{c}Hc@{}}
\hline
Condition & Long\nodrop & Goal\nodrop & Object\nodrop & Spatial\nodrop & Overall Avg.\nodrop \\
\hline
Normal & 98.0\nodrop & 97.0\nodrop & 98.0\nodrop & 95.0\nodrop & 97.0\nodrop \\
Recursive Self-Pred. & 22.0\drop{76.0} & 81.0\drop{16.0} & 52.0\drop{46.0} & 60.0\drop{35.0} & 53.8\drop{43.2} \\
\hline
\end{tabular}
\end{table}
\begin{figure*}[t]
  \centering
  \includegraphics[width=\textwidth]{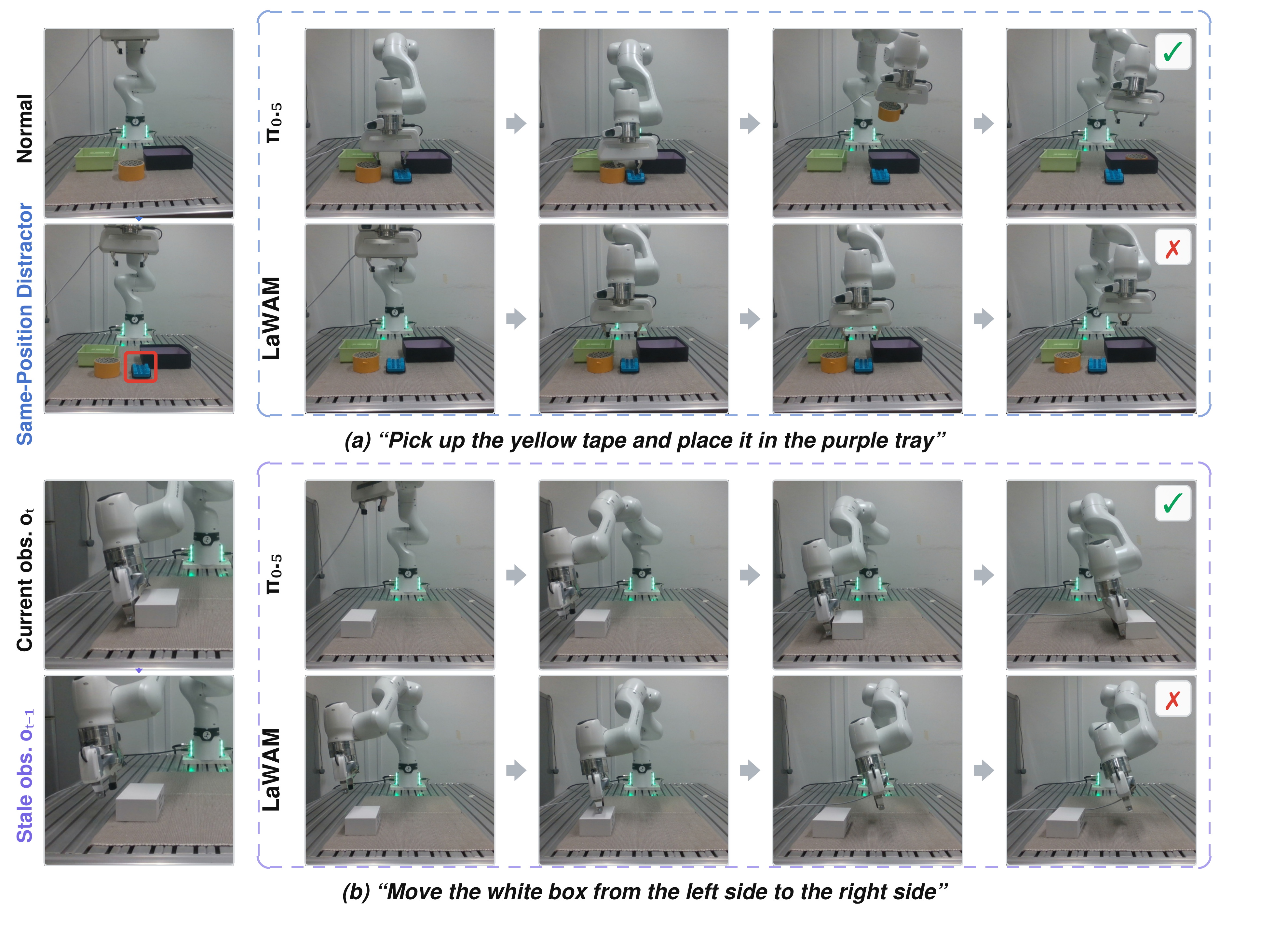}
  \caption{Representative real-world rollouts under two LIBERO-VPro perturbations. (a) Same-Position Distractor on the Pick-and-Place task where the target object is moved nearby while a distractor occupies its original location. (b) Third-Person Camera Staleness on the Move task where the third-person observation is delayed by one replanning cycle while the wrist view remains current. The red box marks the distractor. \textcolor{green!50!black}{\checkmark} = success, \textcolor{red!70!black}{\ding{55}} = failure.}
  \label{fig:real_world_pick_cases}
\end{figure*}
\begin{table}[t]
  \newcommand{\rwbase}[1]{\makebox[2.2em][r]{#1}\nodrop}
  \newcommand{\rwscore}[2]{\makebox[2.2em][r]{#1}\drop{#2}}
  \caption{Success rate (\%) in real-world experiments under Normal and visually perturbed conditions.}
  \label{tab:real_world_results}
  \centering
  \scriptsize
  \setlength{\tabcolsep}{2pt}
  \renewcommand{\arraystretch}{1.08}
  \resizebox{\columnwidth}{!}{%
  \begin{tabular}{@{}cHlHcH*{4}{c}Hc@{}}
    \hline
    \multirow{3}{*}{Task} &
    \multirow{3}{*}{Model} &
    \multirow{3}{*}{Normal} &
    View & Cross-View & \multirow{3}{*}{\shortstack[c]{Same-Pos.\\Distractor}} & Camera &
    \multirow{3}{*}{\shortstack[c]{Overall\\Avg.}} \\
    & & & Unavail. & Contrad. & & Staleness & \\
    & & & (Third) & (Third) & & (Third) & \\
    \hline
    \multirow{2}{*}{\shortstack[c]{Pick-and-\\Place}}
      & $\pi_{0.5}$ & \rwbase{100} & \rwscore{0}{100} & \rwscore{30}{70} & \rwscore{10}{90} & \rwscore{20}{80} & \rwscore{15.0}{85.0} \\
      & LaWAM       & \rwbase{80} & \rwscore{0}{80} & \rwscore{0}{80} & \rwscore{0}{80} & \rwscore{30}{50} & \rwscore{7.5}{72.5} \\
    \hline
    \multirow{2}{*}{Move}
      & $\pi_{0.5}$ & \rwbase{100} & \rwscore{10}{90} & \rwscore{0}{100} & \rwscore{0}{100} & \rwscore{100}{0} & \rwscore{27.5}{72.5} \\
      & LaWAM       & \rwbase{80} & \rwscore{0}{80} & \rwscore{0}{80} & \rwscore{0}{80} & \rwscore{40}{40} & \rwscore{10.0}{70.0} \\
    \hline
  \end{tabular}
  }
\end{table}
As shown in Table~\ref{tab:recursive_suite}, recursive self-prediction causes substantial degradation across all four LIBERO suites, reducing overall success from 97.0\% to 53.8\%. These results show that predicted futures can retain sufficient task-relevant information for partial closed-loop execution, but prediction errors accumulate when generated observations are recursively fed back into the policy. The particularly large degradation on LIBERO-Long is consistent with the greater difficulty of maintaining reliable visual dynamics over longer manipulation sequences.

\subsection{Real-World Evaluation}
\label{sec:real_world}
\textit{Experimental Setup.}
We further examine whether the failure modes identified in simulation transfer to physical manipulation. Experiments are conducted on a Franka Research 3 equipped with third-person and wrist RGBD cameras, both of which are Intel RealSense 435IF. We evaluate $\pi_{0.5}$ and LaWAM as representative VLA and WAM policies on two tasks with different temporal structures. The \textit{Pick-and-Place} task requires a sequence of grasping, transport, and placement actions, whereas the \textit{Move} task relocates a target object through fewer manipulation stages. For each task, we evaluate the Normal condition and four perturbations representative of LIBERO-VPro’s four dimensions: \textit{View Unavailability}, \textit{Cross-View Contradiction}, \textit{Same-Position Distractor}, and \textit{Camera Staleness}. For \textit{View Unavailability}, the third-person observation is replaced with a black frame. For \textit{Cross-View Contradiction}, the target is hidden from the third-person view while remaining visible in the wrist view. For \textit{Same-Position Distractor}, the target object is moved nearby and a distractor is placed at its original position. For \textit{Camera Staleness}, the third-person stream is delayed by one replanning cycle while the wrist observation remains current. The task instruction remains unchanged across conditions. Each model is evaluated for ten trials under every task-condition pair, yielding 200 physical rollouts in total.

\textit{Results.}
As shown in Table~\ref{tab:real_world_results}, both policies perform well under Normal observations but degrade substantially under most visual perturbations. View Unavailability causes near-total failure, confirming the strong dependence on complete multi-view observations observed in simulation. Cross-View Contradiction is similarly disruptive: even when the target remains visible from the wrist camera, inconsistent evidence across views substantially reduces task success. The Same-Position Distractor also causes severe degradation, supporting the simulation finding that policies often rely on familiar spatial layouts rather than consistently re-grounding the target from current visual evidence. Nevertheless, Camera Staleness exhibits stronger task dependence. For Pick-and-Place, delaying the third-person stream sharply reduces the performance of both models, consistent with the need for temporally aligned feedback across multiple manipulation stages. In contrast, the shorter Move task is considerably more tolerant: $\pi_{0.5}$ retains its Normal performance at 100\%, whereas LaWAM drops from 80\% to 40\%. This contrast indicates that the effect of stale observations depends not only on delay magnitude, but also on the temporal structure of the task and the policy’s reliance on each camera stream.

Fig.~\ref{fig:real_world_pick_cases} illustrates representative behaviors. Under the Same-Position Distractor, $\pi_{0.5}$ successfully re-localizes the displaced target in one rollout, whereas LaWAM misses the object. Under Third-Person Camera Staleness, $\pi_{0.5}$ completes the Move task despite the delayed observation, while LaWAM follows a pushing trajectory without establishing contact. Overall, these physical experiments are consistent with the main simulation trends, particularly the sensitivity to missing views, cross-view inconsistency, and violations of learned spatial priors, while also showing that temporal robustness can depend on task structure and model design.

\section{Conclusion}
\label{sec:conclusion}
We have presented LIBERO-VPro, a benchmark for evaluating the closed-loop visual robustness of robotic foundation models. Our results reveal substantial reliance on learned visuospatial priors and vulnerability to disrupted interaction cues, stale observations, and scene changes requiring behavioral adaptation. VLAs and WAMs further exhibit distinct robustness profiles, showing that visual robustness is multi-dimensional and architecture-dependent. We hope LIBERO-VPro provides a systematic diagnostic platform for developing more robust and reliable robotic foundation models.

\bibliographystyle{IEEEtran}
\bibliography{references}
\end{document}